\documentclass[10pt, a4paper]{article}
\usepackage{lrec2022} 
\usepackage{multibib}
\newcites{languageresource}{Language Resources}
\usepackage{graphicx}
\usepackage{tabularx}
\usepackage{soul}
\usepackage{stfloats}
\usepackage{titlesec}
\titleformat{\section}{\normalfont\large\bf\center}{\thesection.}{1em}{}
\titleformat{\section}{\normalfont\large\bfseries\center}{\thesection.}{1em}{}
\titleformat{\subsection}{\normalfont\SmallTitleFont\bfseries\raggedright}{\thesubsection.}{1em}{}
\titleformat{\subsubsection}{\normalfont\normalsize\bfseries\raggedright}{\thesubsubsection.}{1em}{}
\renewcommand\thesection{\arabic{section}}
\renewcommand\thesubsection{\thesection.\arabic{subsection}}
\renewcommand\thesubsubsection{\thesubsection.\arabic{subsubsection}}

\usepackage{epstopdf}
\usepackage[utf8]{inputenc}

\usepackage{hyperref}
\usepackage{xstring}

\usepackage{color}

\usepackage{microtype}
\usepackage{times}
\usepackage{latexsym}
\usepackage{float}
\restylefloat{table}
\usepackage{booktabs}
\usepackage{times}
\usepackage{latexsym}
\usepackage{multirow}
\usepackage{graphicx}
\usepackage{multicol}
\usepackage{amsmath}
\usepackage{amsfonts}
\usepackage{amssymb}
\usepackage{subfig}
\usepackage{wasysym}
\usepackage[nameinlink]{cleveref}
\usepackage{hyperref}
\usepackage{tikz}
\usepackage{numprint}
\usepackage{enumitem}

\title{Human Schema Curation via Causal Association Rule Mining}

\name{\begin{tabular}{c} Noah Weber$^{*,a}$, Anton Belyy$^{*,a}$, Nils Holzenberger$^{*,\gamma,a}$, \\ Rachel Rudinger$^b$, Benjamin Van Durme$^a$ \end{tabular}}

\address{
\begin{tabular}{c c}
$^a$Johns Hopkins University & $^b$University of Maryland, College Park\\
3400 N. Charles Street & 8125 Paint Branch Drive \\
Baltimore, Maryland, USA & College Park, Maryland, USA \\
\end{tabular} \\
\begin{tabular}{cc}
\{nweber6, abel, nilsh, vandurme\}@jhu.edu & rudinger@umd.edu
\end{tabular}
}

\abstract{
Event schemas are structured knowledge sources defining typical real-world scenarios (e.g., \emph{going to an airport}). We present a framework for efficient human-in-the-loop construction of a \emph{schema library}, based on a novel script induction system and a well-crafted interface that allows non-experts to ``program'' complex event structures. Associated with this work we release a schema library: a machine readable resource of 232 detailed event schemas, each of which describe a distinct typical scenario in terms of its relevant sub-event structure (\emph{what} happens in the scenario), participants (\emph{who} plays a role in the scenario), fine-grained typing of each participant, and the implied relational constraints between them. We make our schema library and the SchemaBlocks interface available online.${}^1$${}^2$
 \\ \newline \Keywords{schemas, script induction, dataset curation, annotation interfaces} }

\begin{document}

\maketitleabstract
\let\svthefootnote\thefootnote
\let\thefootnote\relax\footnotetext{$^*$Equal contribution. Order decided via wheel.}
\footnotetext{$^\gamma$Corresponding author}
\let\thefootnote\svthefootnote
\stepcounter{footnote}\footnotetext{Schema library:\\\url{https://nlp.jhu.edu/schemas/schemas.zip}}
\stepcounter{footnote}\footnotetext{Interface: \url{https://nlp.jhu.edu/demos/sb}} 

\section{Introduction}

What is implied by the invocation of a real-world scenario such as, say, a \emph{criminal trial}? 
From one's knowledge of the world, one makes a myriad of inferences: the scenario typically starts with the \emph{defendant} being accused and brought to court, it likely contains events such as the presentation of evidence by a \emph{prosecutor}, and it ends with the \emph{judge} announcing the final verdict.

This type of scenario-level knowledge is recognized as being vital for text understanding \cite{sch77scripts,minsky74,bower1979scripts,abbott1985representation}: scripts can help with coreference resolution, disambiguating word meaning, and making inferences \cite{lehnert83boris}. However, explicitly annotating this knowledge in a way useful to language processing systems has proven to be a difficult task.
At one end, one may try to hand-engineer this knowledge in a richly detailed format \cite{dejong1983acquiring,mooney,mueller1999database}. While this facilitates precise inferences, it requires an onerous annotation effort carried out by experts, and hence tends to be difficult to scale. On the other end, one may employ data-driven methods to automatically induce this knowledge \cite{chambers:09,balasubramanian2013generating,Rudinger2015}, at the price of noise and a severe loss of detail.
\newcite{descript} take a semi-automatic approach, taking advantage of both automatic and annotator-driven components. The authors use an initial human annotation to obtain high quality event sequence descriptions for a target scenario, before using semi-supervised clustering to aggregate these annotations \cite{descript_clusters,regneri2010learning}.

\begin{figure}[t]
\centering
\includegraphics[width=1\columnwidth]{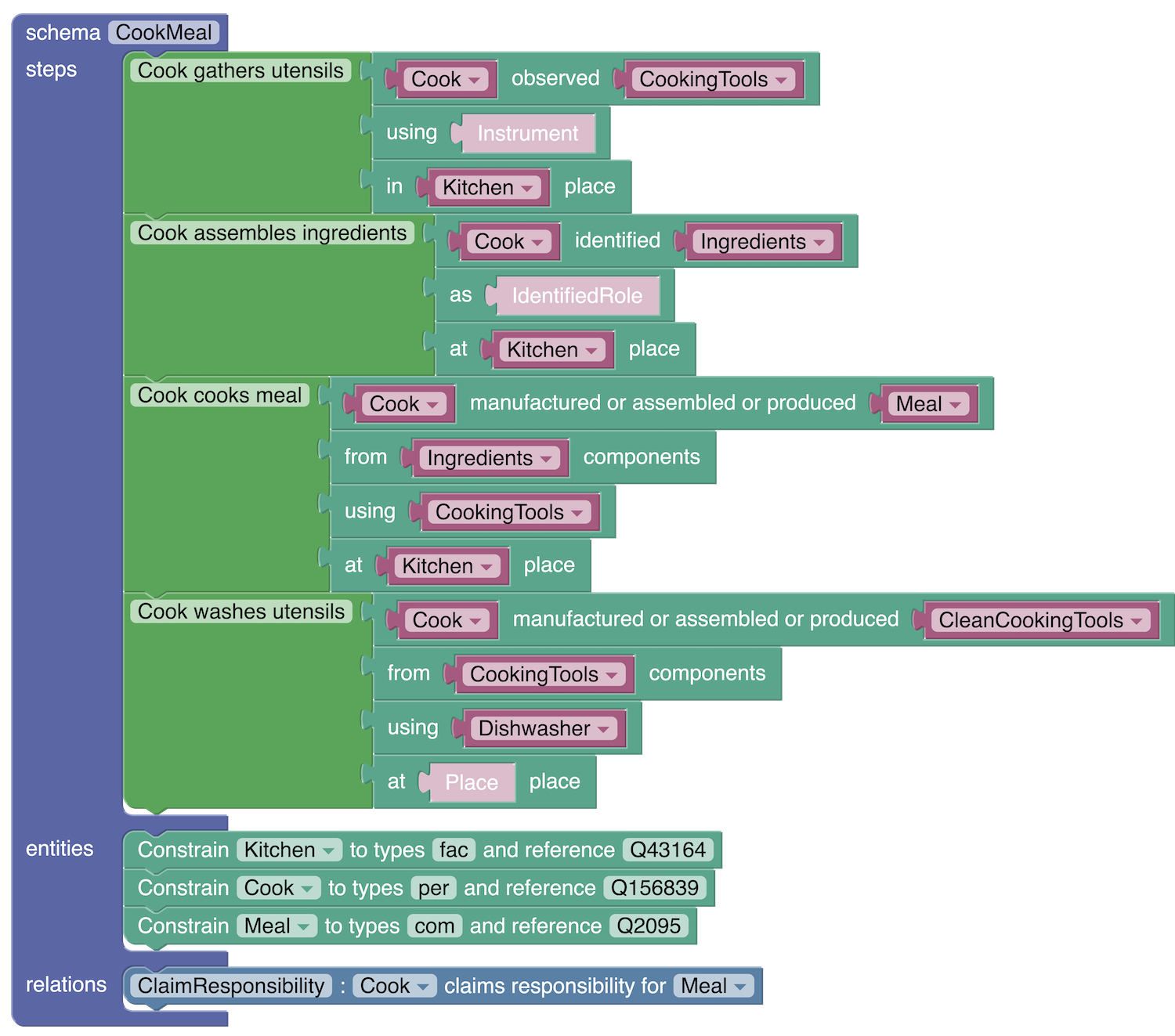}
\caption{An example event schema from our library, induced from a skeleton mined by Causal Association Rule Mining (\Cref{sec:auto}) and fully fleshed out by an annotator using our SchemaBlocks interface (\Cref{sec:manual}).}
\label{fig:example}
\end{figure}

In this paper, we also adopt a semi-automatic approach in order to facilitate the creation of a new annotated resource of structured, machine readable \emph{event schemas}. As depicted in \Cref{fig:example}, each event schema characterizes a real-world scenario, describing the \emph{events} the scenario typically involves, the \emph{participants} of these events, their role and typing information, and the implied \emph{relations} between these participants. Our workflow follows two main steps. First, we automatically induce what we term as \emph{skeleton schemas}: argumentless event sequences that form the outline of an event schema. Second, using our SchemaBlocks interface, we have human annotators ``flesh out'' the manually selected skeleton schemas by adding argument, role, typing, and relational information, in addition to a name and description of the scenario the schema describes.

The main contributions of this paper are:\footnote{Resources tied to this paper are grouped here:\\ \url{https://nlp.jhu.edu/schemas}}

\begin{enumerate}
    \item a new semi-automatic script induction system, which combines two recent advances in automatic script induction \cite{belyy-van-durme-2020-script,weber-etal-2020-causal} with a novel SchemaBlocks annotation interface, to elicit common sense knowledge from crowdworkers,
    \item a resource of 232 schemas, 150 of which are semi-automatically induced, with the rest created manually from textual descriptions, and
    \item two novel evaluation metrics for schemas: \emph{corpus coverage}, an automatic metric which computes coverage of schemas on a text corpus, and \emph{schema intrusion}, a human-based metric which quantifies the coherence of each schema, similarly to the word intrusion task~\cite{chang09reading}.
\end{enumerate}
\section{The Anatomy of a Schema}
Conceptualizations of what constitutes a \emph{schema} differ across the literature. A schema in our resource is constructed from three basic elements:

\begin{enumerate}
    \item events,
    \item its \emph{participants}, which are the entities that participate in these events, and
    \item the relations between these participants.
\end{enumerate}

The atomic types of events, entities, and relations are defined by the DARPA KAIROS Phase~1 (v3.0) ontology.\footnote{The ontology can be downloaded here:\\\url{https://nlp.jhu.edu/schemas/ont.xlsx}} It consists of 67 event types, 24 coarse-grained entity types, and 46 relation types.
The KAIROS ontology was selected because this work was carried out in the context of a larger effort, where collaborators used schemas for information extraction. While that choice influenced the content of the schemas produced here, our methods are ontology-agnostic, and our interface's building blocks (see \Cref{sec:schemablocks}) could easily be adjusted to elicit schemas from humans with any type of ontology, including more general and more flexible ontologies such as FrameNet \cite{baker1998berkeley}.

\paragraph*{Events}
In this work, the backbone for the meaning of a schema is the temporally ordered chain of events that it describes. The individual events that make up this chain are drawn from a taxonomy of event types (e.g., an \emph{Acquit} event, a \emph{Transportation} event). In addition, each event type has specific participants (e.g., the \emph{Defendant} or \emph{Transporter}), to be linked to entities. While we use the term ``chain'' to describe the sequence of events defined in a schema, the schemas presented here need not always be ordered as a linear chain. In our schemas, subsequences of events may be marked either as occurring in a linear temporal order, in an arbitrary temporal order, or as forming mutually exclusive branches.

\paragraph*{Participants}
Participants fill the roles specified by each event in the schema. The same participant can (and usually will) be used to fill different roles across different events, indicating a co-referring relationship. All participants may also take on types: either coarse grained types defined in the KAIROS ontology (including types such as \emph{person}, \emph{organization}, \emph{commercial item}, etc), or fine grained types defined as references to Wikidata: for instance, on \Cref{fig:example}, Q156839 refers to a Wikidata entity for ``cook'', which substantially narrows down a more generic type \textit{person}. Our annotated schemas utilize both KAIROS and Wikidata types.

\paragraph*{Relations}
Relations between participating entities are the last ingredient of the schemas defined here. These relations are also drawn from the KAIROS ontology. As of now, all relations are defined between two entities, each of which participate in at least one event defined in the schema: e.g. {ClaimResponsibility}(\textit{Cook}, \textit{Meal}) in the ``CookMeal'' schema on \Cref{fig:example}.

\section{Induction of Skeleton Schemas} \label{sec:auto}
Our system first automatically induces what we term as \textit{skeleton schemas}: argumentless event sequences forming an outline of a potential event schema. A selected group of these skeleton schemas is then passed to annotators to manually flesh out the full event schemas.  By starting the schema creation with an automatic, data-driven step, we allow the data to ``speak for itself'' with regards to what kinds of topics and scenarios we might want to target given a specified domain. The fact that the base of the schemas has some connection to our targeted domain gives at least some assurance that the final schemas will be applicable towards making commonsense inferences when used in real-world applications.

The automatic system for skeleton schema induction combines two recent advances in schema induction:

\begin{enumerate}
    \item an Association Rule Mining (ARM) based algorithm presented in \newcite{belyy-van-durme-2020-script}, which efficiently finds all event subsequences with sufficient support in the data, and
    \item a script compatibility scoring model presented in \newcite{weber-etal-2020-causal}, which finds high quality subsequences output by the ARM method, and combines them  to form full skeleton schemas.
\end{enumerate}

We give a brief overview of each of these approaches and how they are used in our system below. 

\subsection{Mining Associations for Script Induction} \label{sec:arm}
\newcite{belyy-van-durme-2020-script} show how prior classic work in automatic script induction (primarily the line of work following \newcite{chambers:08}) can be better recast as a problem of Association Rule Mining. ARM works with a dataset where each datapoint is a set of items. In the script induction setting, an \emph{item} is an event, and a datapoint is the set of events appearing in a document and sharing some co-referring argument. The ARM approach consists of two distinct stages:

\begin{enumerate}
    \item \textbf{Frequent Itemset Mining}. This step searches for subsequences of events which have enough support in the dataset. What is considered ``enough'' is defined by a user-set hyperparameter. To do this efficiently, \newcite{belyy-van-durme-2020-script} make use of the FP-growth algorithm \cite{han2000mining}.
    \item \textbf{Rule Mining}. This step uses the frequent itemsets mined from the previous step in order to define rules in a form similar to Horn clauses.
\end{enumerate}

In our system, we make use of only step~1 of the process defined above, mining event subsequences which have enough support in our targeted domain data. We mine event subsequences from the NYTimes portion of Gigaword \cite{graff2003english}. The output of this step is a large set of potentially interesting event subsequences. 

\subsection{Building Schemas with a Causal Scorer}
\label{sec:building-schemas}

The step presented in the previous section leaves us with a fairly large inventory of event subsequences, not all of which may be useful or relevant for the creation of schemas. There are, hence, two problems at hand:

\begin{enumerate}
    \item how to filter out lower quality subsequences, and
    \item how to create skeleton schemas from the filtered inventory of event subsequences.
\end{enumerate}

Both problems are handled via the causal inference based scoring approach of \newcite{weber-etal-2020-causal}. This approach  defines a scoring function, $\text{cscore}(\cdot, \cdot)$ which, taking in two events $e_1$ and $e_2$, outputs a score proportional to the aptness of $e_2$ following $e_1$ in a script. As an example, ``trip'' and ``fall'' should take on high scores, while ``trip'' and ``eat'' should not. The approach builds upon reasonable assumptions on the data generation process to overcome conceptual weaknesses in prior approaches, and was shown to output scores more in line with human judgments of script knowledge. We refer readers to the paper for details. 

In order to create our skeleton schemas, we first use the trained scoring module from \newcite{weber-etal-2020-causal}, which was trained on the Toronto Book corpus \cite{zhu15aligning,kiros15skip}, to score all subsequences obtained via the process described in \Cref{sec:arm}. Since the causal scoring module is only defined pairwise, we take the following average as the assigned score for a subsequence ${S = (e_1,...,e_N)}$ of length $N$:
\[
\text{score}(S) = \frac{2}{N(N-1)}\sum^{N-1}_{i=1} \sum^{N}_{j=i+1} \text{cscore}(e_i, e_j)
\]

We take the top $T$ of these subsequences. To ensure that a diverse set of events are selected in the subsequences, we remove those in which all event types in the sequence have been used at least 50 times by higher scoring subsequences. 

The score function above is biased towards shorter subsequences: picking the highest scoring pair of events in a subsequence creates a higher-scoring subsequence. To mitigate this, our final step involves joining together subsequences to create larger chains. For each of these $T$ subsequences, we find the highest scoring event that may be appended to the subsequence. We then find other subsequences that start with this event, and append the highest scoring one to the existing subsequence.

The top $C$ of these larger subsequences are then given to a curator (one of the authors), who manually selects chains to be passed to human annotators as skeleton schemas. This is done as an expedient to ensure both the diversity and quality of the resulting schema annotations. We pick ${C=\npthousandsep{,}\numprint{1000}}$ as the upper feasible limit for a manual curator. To make sure there are enough potential merges, we set ${T = 100C}$. Finally, our annotation budget was enough to turn the top $150$ of these $C$ chains into schemas (see \Cref{sec:flesh_out}).
\section{Annotation with SchemaBlocks}\label{sec:manual}
After skeleton schemas are induced, we want to include rich commonsense information (i.e. event participants, their types and relations) in addition to the event sequence. As the existing induction tools struggle to induce these fully automatically, we involve a human in this process. We describe the newly proposed schema annotation interface, SchemaBlocks, and show how it can be used to 

\begin{enumerate}
    \item create schemas from scratch (\Cref{sec:manual_ann}), and
    \item flesh out skeleton schemas (\Cref{sec:flesh_out}).
\end{enumerate}

We also share our annotation guide and some relevant statistics on the annotation process.

\subsection{SchemaBlocks Annotation Interface}
\label{sec:schemablocks}

\begin{figure}[t]
\includegraphics[width=1\columnwidth]{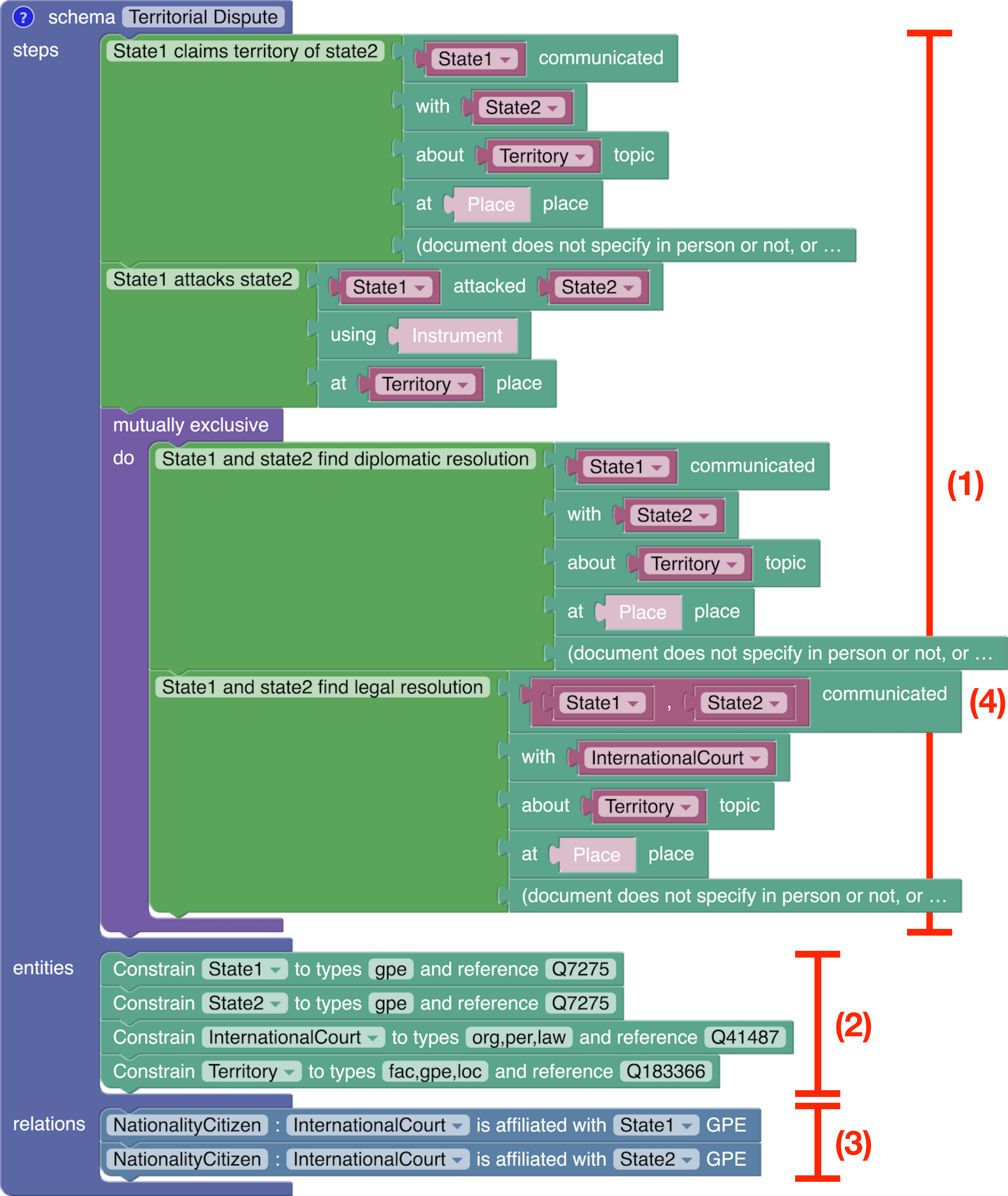}
\caption{An excerpt from one of the released schemas, featuring: (1)~mutually exclusive events, (2)~entity types, (3)~entity relations, and (4)~a slot filled with more than one entity (\textit{State1} and \textit{State2}), reflecting that an event may include multiple participants under the same role. Participants left in light pink by the user are defined as part of the event type in KAIROS, but not instantiated (reified) in the event schema.}
\label{fig:schema}
\end{figure}

\begin{figure}[t]
\includegraphics[width=1\columnwidth]{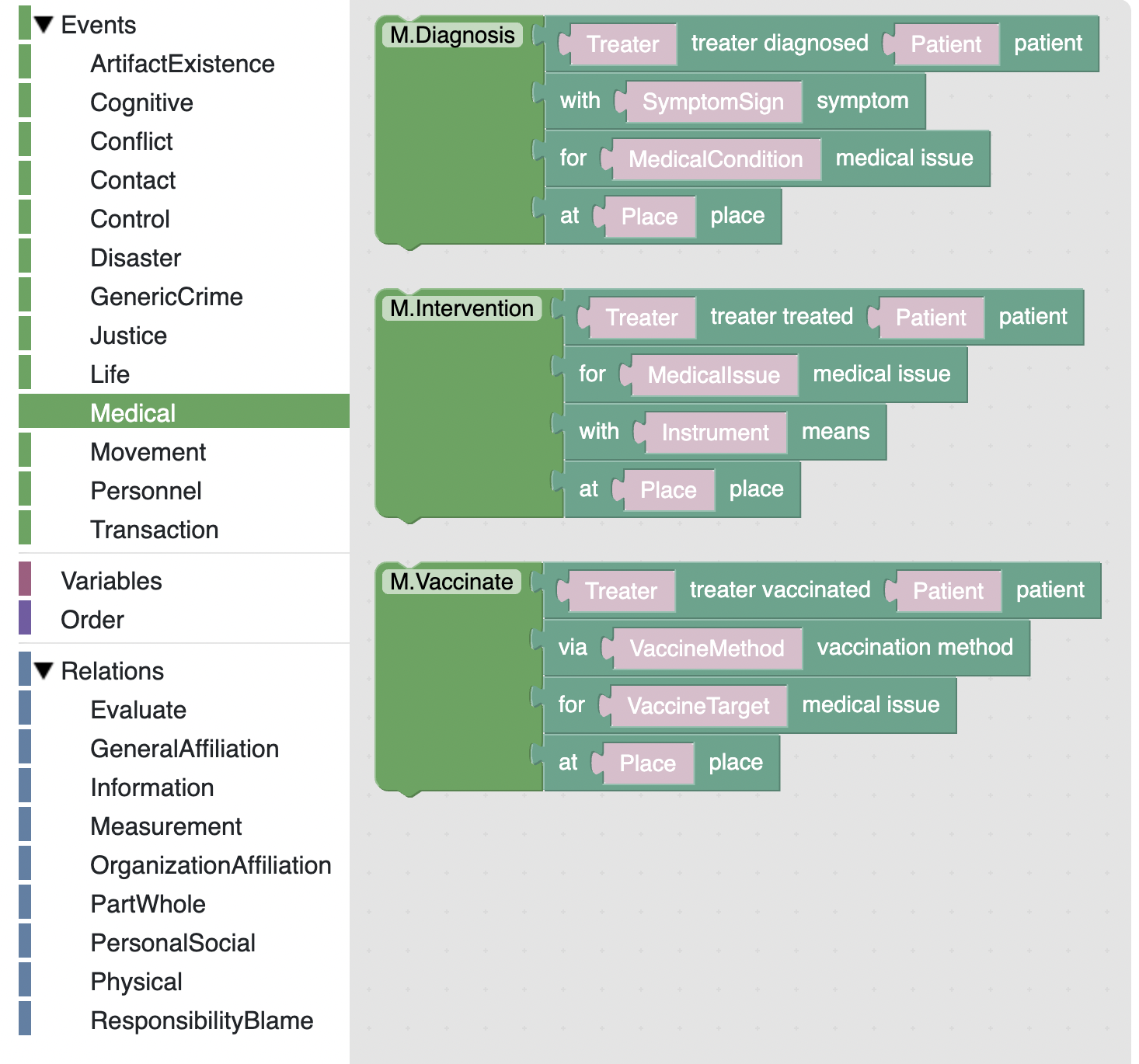}
\caption{SchemaBlocks dashboard displaying high-level event and relation types from the KAIROS ontology. The ``Medical'' event category is further expanded to show subtypes. ``Variables'' and ``Order'' blocks allow to assign multiple entities to a single participant of an event, and to specify the ordering of events, respectively.}
\label{fig:dashboard}
\end{figure}

SchemaBlocks is a Web-based tool\footnote{Source code of SchemaBlocks:\\\url{https://github.com/AVBelyy/SchemaBlocks}}
that provides a way to display and  modify the contents of a schema by representing its units -- events and arguments, entity relations and types -- as \textit{blocks}, that can be stacked and nested. An example schema is shown in \autoref{fig:schema}. In addition to capturing schema events, participants, and their relations, the interface also allows for the representation of entity coreference, event ordering, and the mutual exclusivity of events.

To get started, an annotator needs to become familiar with the ontology, which defines the vocabulary of blocks used to build schemas. In the interface, this is displayed as the dashboard, organized hierarchically for convenience. \autoref{fig:dashboard} shows all levels of the ontology hierarchy for the ``Medical'' event category.
The block interface is flexible and could be adapted to a similar event ontology, such as FrameNet \cite{baker1998berkeley}, ACE \cite{doddington2004automatic} and ERE \cite{song2015light}. For larger event ontologies, it may be helpful to implement search functionalities into the interface to facilitate quicker access to a specific event in the ontology. The core annotation process with SchemaBlocks would, however, remain the same. Such features may be worthwhile additions in future versions of SchemaBlocks.

SchemaBlocks' interface is primarily based on the Google Blockly library.\footnote{\url{https://github.com/google/blockly}} On top of the UI primitives provided by Blockly, we implement ontology-to-blocks and blocks-to-JSON converters. This allows to transform a structured ontology description into the set of Blockly blocks, which the user can manipulate to create a schema, and when they are done, transform their block-based schemas into a machine-readable format. During schema creation, we also continuously run type checking and type inference over schema entities, so that if a user breaks ontological type constraints, they will be notified and the relevant entity blocks will be highlighted until the error is fixed.

Our choice of block-based representation is inspired by Scratch \cite{resnick2009scratch}, a prominent tool that engages children to learn the basics of computer programming. By enabling users to program schemas using ontology-specific blocks --- as opposed to general-purpose text formats such as JSON or XML --- we were able to engage annotators with non-programming backgrounds and annotate schemas at a faster rate. The annotators in our study (undergraduate students with non-CS majors) found the interface easy-to-use and left overall positive feedback. To familiarize annotators with the interface, we provided them with a guide prior to running the annotation: \url{https://nlp.jhu.edu/schemas/guide.pdf}.

\subsection{Annotating Schemas from Scratch} \label{sec:manual_ann}
In the first annotation round, annotators were provided with 82 textual descriptions of schemas from the KAIROS Schema Learning Corpus (LDC2020E25). This corpus contains textual definitions for 82 \emph{complex events} (CEs), which we aim to transform into event schemas. Each complex event is given a title, a 2-3 sentence description, specifications of the scope of the complex event (i.e., when and where the complex event should be considered initiated or finished), and the series of steps that defines the complex event. Each step is defined with a title specifying the event type of the step, a short one sentence description, and expected high-level event types that may happen as subevents.\footnote{All of this in natural language; no event ontology is used.}

The annotators are then tasked with translating these textual descriptions of schemas into a machine readable form via our SchemaBlocks interface. Relations and entity types are not specified in the textual descriptions, so annotators are instructed to annotate for relations that must be true throughout all steps of the schemas, as well as provide coarse- and fine-grained types. Annotators reported an average time of 30 minutes to annotate a CE into a schema, with 82 schemas being the product of this annotation task. The number of events in each of 82 schemas ranges from 2 to 10, with 6 being the median.

\subsection{Fleshing out Skeleton Schemas}\label{sec:flesh_out}
In the second annotation round, annotators were asked to ``flesh out'' the skeleton schemas from \Cref{sec:auto}, into fully-fledged schemas. Given a skeleton schema, we import it into SchemaBlocks as a partially filled out schema, where only events are specified. We then present these partially filled out schemas to annotators and task them with determining:
\begin{itemize}
    \item What scenario the partially filled out schema is describing. This includes naming the schema, as well as writing a brief textual description on what it is about. 
    \item Who the participants of the given events are, what types (coarse- and fine-grained) they take on, and which roles are filled with co-referring participants.
    \item What relations hold between the above defined entities. The criteria for annotating relations here is the same as before.
\end{itemize}
Given that this annotation is designed to be similar to the one presented in \Cref{sec:manual_ann}, all annotators who participated in the first annotation effort required little extra training to complete this annotation (only a single one-hour training session). Again, annotators reported around a 30-minute average to annotate a schema. The end result of this fleshing out process is an additional 150 schemas.  The number of events in this additional set ranges from 3 to 6, with 4 being the median.
\newcommand{\myspace}{\hspace{5pt}}

\section{Schema Library Evaluation} \label{sec:eval}
In this section, we evaluate our schema library\footnote{At the time of writing, there were no other publicly available schema libraries using the KAIROS ontology, which limited the cross-library comparisons we could run.}, looking at schemas' internal coherence as well as usefulness of schemas for downstream tasks. Namely, we evaluate the coherence of the event sequence in a schema by measuring the accuracy on the \emph{schema intrusion} task (\Cref{sec:intruder-task}). Then, we compute how many documents in a given corpus are ``covered'' by the schema library as a whole, using the \emph{corpus coverage} metric (\Cref{sec:corpus-coverage-eval}). Finally, we report the results on several ranking tasks, using event schemas as structured queries to rank multimodal documents, and vice versa (\Cref{app:ranking_eval}).

We evaluate both the library of schemas created from scratch (\Cref{sec:manual_ann}, ``82 schemas''), as well as the library created from schema skeletons (\Cref{sec:flesh_out}, ``150 schemas''). The two methods used to obtain each library are not meant to be compared directly, because they are two different ways of eliciting schemas from humans. Each method relies on a different starting point for schemas: respectively, textual descriptions of schemas, and event chains induced from a corpus. Choosing which one to use depends on the resources~available.

\subsection{Evaluation Datasets} \label{sec:datasets-eval}

\paragraph*{Gigaword} We pick a random subset of 100K documents from the NYTimes portion of the Fifth Edition of the English Gigaword corpus \cite{graff2003english}, spanning the New York Times news articles from years 1994--2010. We use this corpus for corpus-based evaluation in the schema intrusion task (\Cref{sec:intruder-task}), as well as to compute corpus coverage (\Cref{sec:corpus-coverage-eval}). 

\paragraph*{CC-News} We pick a subset of 300K news articles from English, Russian, and Chinese CC-News \cite{nagel2016cc}. To do that, we perform language ID over the original CC-news collection, using the cld3 library along with the ``meta\_lang'' field from a particular news source.
We then take a random subset of 100K documents for each language to evaluate corpus coverage (\Cref{sec:corpus-coverage-eval}) in a cross-lingual scenario.

\paragraph*{KAIROS SLC}
As part of the KAIROS Schema Learning Corpus (SLC), the Linguistic Data Consortium (LDC) has annotated 924 multilingual multimodal documents (covering images, audio, video, and text in English and Spanish) with KAIROS event types, labeling each document with one of 82 complex events mentioned earlier in \Cref{sec:manual_ann}.\footnote{At the time of writing, these annotations have been split into three collections: LDC2020E24, LDC2020E31, and LDC2020E35. While rarely freely released, historically, such collections are eventually made available under a license to anyone, under some timeline established within a program.} The CE label indicates the complex event (from LDC2020E25) that best applies to a document. Each CE label is covered by 11 documents on average, one label per document. Out of 924 documents, 921 have partial event-only annotations and 36 have complete annotations (with identified and provenance linked entities and relations). Given the sparsity of complete annotations, we use the event-only annotated documents in order to compute ranking-based metrics (\Cref{app:ranking_eval}).

\subsection{Schema Intrusion Task}\label{sec:intruder-task}

\begin{figure*}[t!]
\centering
\includegraphics[width=\textwidth]{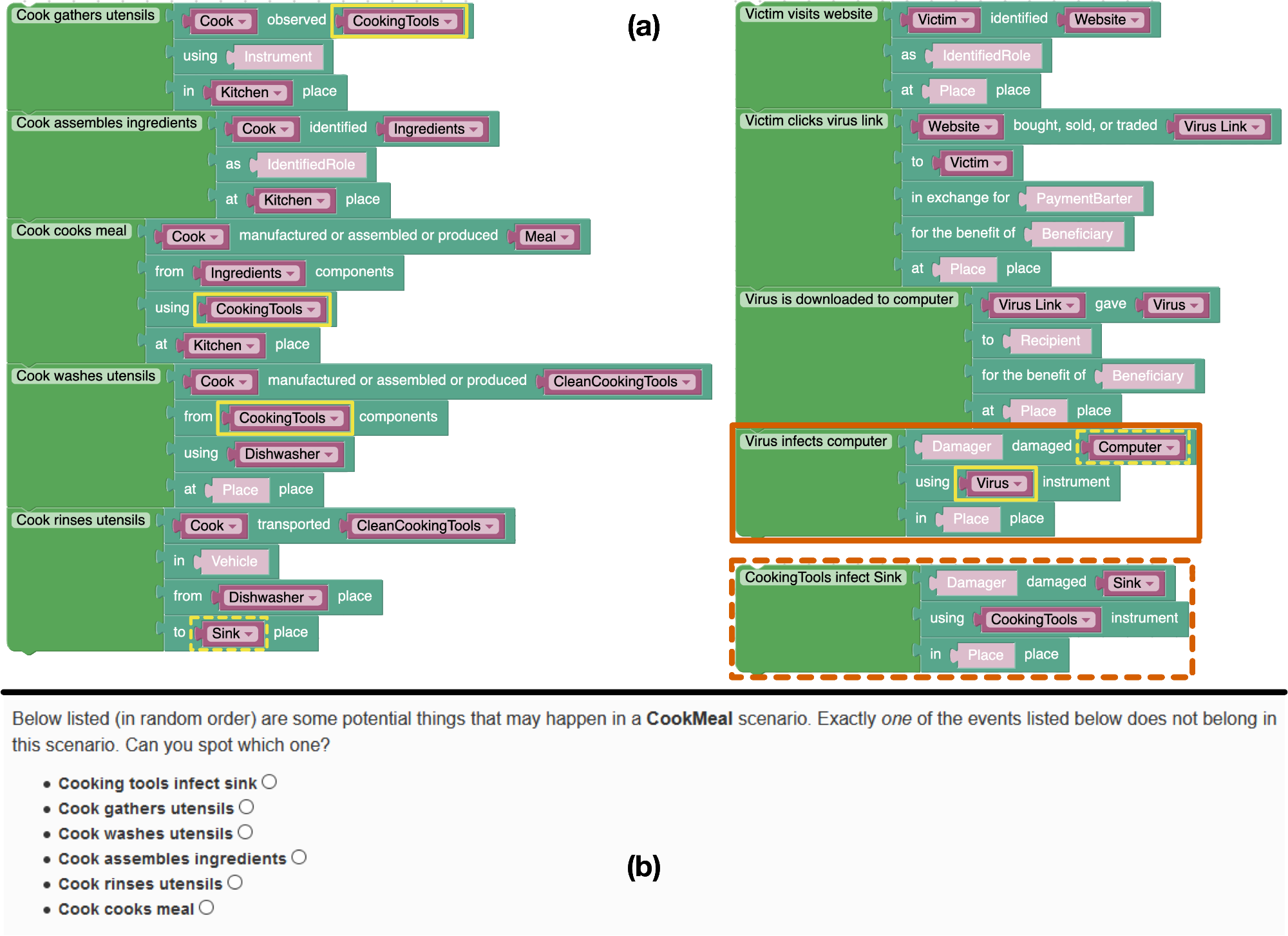}
\caption{Example schema intrusion sample. (a)~A step from the ``Download Computer Virus'' schema (right) is added to the ``Cook Meal'' schema (left). The step ``Virus infects computer'' (solid orange) is sampled, with ``Computer'' replaced with ``Sink'' (dashed yellow) and ``Virus'' replaced with ``CookingTools'' (solid yellow). This yields the intruder ``Cooking Tools infect Sink'' (dashed orange). (b)~The schema with the intruder is presented to the human annotators.}
\label{fig:intruder}
\end{figure*}

To measure to what extent our schemas form meaningful units, and how much the content of one schema overlaps with that of another, we introduce the \emph{schema intrusion} task. Schema intrusion is similar in spirit to \emph{word intrusion} for topic models \cite{chang09reading}. At a high level, for each schema $S$ from our library, we pick a step from a different schema $S'$ and add it to $S$. We present $S$ to a human evaluator with the task of picking the intruder. The more coherent and exhaustive each schema is, the more the intruder should stick out as being out of place, untypical, or at least irrelevant.

Simply inserting a step from schema $S'$ into schema $S$ gives rise to artefacts, making it easy to spot the intruder without reasoning about the coherence of schema $S$. Consider \Cref{fig:intruder}(a): inserting a step from 
``Download Computer Virus'' into ``Cook Meal'' would introduce the participant ``Computer'' or ``Virus'', which gives the step away as the intruder, regardless of schema coherence. Thus, we need a way of renaming participants of the step we pick from ``Cook Meal'' before inserting it into ``Download Computer Virus''. To avoid any bias from the ordering of the steps, we shuffle the steps before showing them to the annotator.

Building instances of the intrusion task to present to annotators is a sampling procedure. In the following, we detail two ways to define the samples and their unnormalized weights: a \emph{library-based} method and a \emph{corpus-based} method. To sample an intruder for schema $S$ with the library-based method, we need to sample a step $e$ from a schema $T$, as well as a mapping ${M = \{x \rightarrow y\}}$ of the participants of $e$ to participants of $S$. The mapping $M$ is used to rename participants from $e$ with names that come from $S$, camouflaging $e$'s participants to look like participants from $S$, and mitigating the artifacts mentioned above. A sample is a tuple ${(T, e, M)}$ with weight $w$. Let ${\mathrm{type}(x)}$ be the set of types associated with participant $x$ under the KAIROS ontology. For instance, in \Cref{fig:intruder}(a), ``Virus'' is associated with the types \{\texttt{abs}, \texttt{com}\}. We use the Jaccard index ${J(A,B) = \frac{|A \cap B|}{|A \cup B|}}$ to measure overlap in types between participants, with ${J(\varnothing,\varnothing) = 0}$. We compute $w$ as the geometric mean of type overlap between participants: ${w = [\prod \limits_{x,y \in M} J(\mathrm{type}(x), \mathrm{type}(y))]^{\frac{1}{|M|}}}$. The use of the geometric mean is meant to exclude type incompatibilities, any of which would set the weight to 0. In addition, we reject any sample which, after renaming $e$'s participants, would create a step already in $S$.

The corpus-based method finds candidate steps $e$ by relying on documents. We first match schemas in the library with documents, in a process called \emph{schema inference}. We describe a document using the events and participants from an ontology. We frame schemas as predicates over tuples of events, relying on that same ontology, and using Horn clauses to capture the relationships between schemas, events and their participants. Using the formalism and tools of Probabilistic Soft Logic \cite{bach2017hinge}, schema inference is recast as a convex optimization problem, and solved. This procedure is further detailed in \Cref{app:schema_inference}. Here, a sample is ${(d,T,e,M)}$ where $d$ is a document such that both $S$ and $T$ match $d$. As part of matching with $d$, some of the participants in $S$ and $T$ will be matched with entities present in $d$. Let $\mathrm{ent}(x,d)$ be the (possibly empty) set of entities associated with participant $x$ in document $d$. For each tuple ${(d,T,e,M)}$, similarly to the weight based on type mentioned above, we compute ${w = [\prod \limits_{x,y \in M} J(\mathrm{ent}(x,d), \mathrm{ent}(y,d))]^{\frac{1}{|M|}}}$. For this corpus-based method, we used the Gigaword corpus mentioned in \Cref{sec:datasets-eval}, only keeping documents that contain between 2 and 10 events, to be comparable to the number of events in our schemas.

The mapping ${M = \{x \rightarrow y\}}$ is used heuristically to modify the description of $e$, by replacing occurrences of string $x$ by string $y$. We manually ensured that intruders would not be given away by artifacts that come up during this procedure, as follows. First, we normalized the form of the step descriptions in all schemas, standardizing verb inflection and syntax. Second, we reviewed each intruder instance and corrected any grammatical inconsistencies introduced by heuristically renaming participants. Finally, human evaluators for the task were only presented with the textual description of steps.
As a result, any difference between schema curators in the use of the KAIROS ontology, the presence or absence of explanatory comments or any other SchemaBlocks feature, cannot have any influence on the human annotator's ability to spot the intruder.
\Cref{fig:intruder}(b) shows one instance of the task, as presented to the annotators.

Each intrusion task, consisting of one schema (as shown in \Cref{fig:intruder}(b)), is completed by \emph{three} separate annotators on the Mechanical Turk platform (see \Cref{app:eval_details}).
Results of this evaluation are shown in \Cref{tbl:intruder}. The total accuracy for the task (``Total''), which considers each annotator separately, is far above the accuracy of randomly picking the intruder (``Random''). This shows that our schemas form units meaningful for humans. In addition, more than 85\% of times, at least one out of the three annotators was able to spot the intruder (``1~Ann.''), far above the corresponding accuracy of random picks (``Random~1''). 
Finally, even when we require all three annotators to agree on the intruder (``3~Ann.''), the accuracy is still far above that of picking at random (``Random~3''). The differences in accuracy between ``$n$~Ann.'' and the corresponding ``Random~$n$'' are significant (p-value~${\ll 0.01}$), as measured by the two-sided McNemar's test.

Contrasting both methods to sample the intruder, it seems both are roughly equally hard to spot. One would expect the corpus intruders to be more difficult to detect, since their sampling is informed by documents. While this is true for the 150 schemas, it is not for the 82 schemas. This can be explained by the fact that some schemas match many documents, while some match fewer documents. Similarly, some schema steps match more documents than others as certain events come up more often than others in the corpus. This likely induces a skew in the types of events that intruders typically cover, which in the case of the 82 schemas, introduces regularities that make the intruders stick out.

\begin{table}[t!]
\centering
\small
\resizebox{0.9\columnwidth}{!}{
\begin{tabular}{l ccc ccc}
\toprule
 \multicolumn{1}{l}{}               & \multicolumn{3}{c}{Library}   & \multicolumn{3}{c}{Corpus} \\ \cmidrule(r){2-4} \cmidrule(r){5-7}
                        & 82    & 150   & 232           & 82    & 150   & 232 \\ \midrule
 Total     & 62.0  & 67.3  & 65.4          & 73.2  & 61.7  & 65.8  \\
 1 Ann.           &  84.1  & 86.7  & 85.8          & 93.0  & 83.3  & 86.6 \\
 2 Ann.           &  64.6  & 71.3  & 69.0          & 79.3  & 62.0  & 68.1 \\
 3 Ann.        & 36.6  & 44.0  & 41.4          & 48.0  & 40.0  & 43.0 \\
 \addlinespace
 Random   & 16.0           & 21.2           & 19.3           & 16.0           & 21.2           & 19.3 \\
 Random 1 & 40.2           & 50.8          & 47.1           & 40.2           & 50.8          & 47.1 \\
 Random 2 & \phantom{0}7.3 & 11.7           & 10.1           & \phantom{0}7.3 & 11.7           & 10.1 \\
 Random 3 & \phantom{0}0.5 & \phantom{0}1.0 & \phantom{0}0.8 & \phantom{0}0.5 & \phantom{0}1.0 & \phantom{0}0.8 \\
\bottomrule
\end{tabular}}
\caption{Human accuracy on the schema intrusion task, as \%. ``82'', ``150'' and ``232'' refer to the size of the schema library used. ``1~Ann.'' (resp. ``2~Ann.'') considers the intruder found if at least 1 (resp. 2) annotator(s) found the intruder. ``3~Ann.'' counts an intruder as found only if it was found by all 3 annotators. ``Total'' considers each vote separately.  ``Random'' is the expected accuracy of picking the intruder at random. ``Random~$n$'' is the expected accuracy of picking three intruders at random with replacement, and having at least $n$ of those be the correct answer.}\label{tbl:intruder}
\end{table}

\subsection{Corpus Coverage}\label{sec:corpus-coverage-eval}

\begin{table*}[!t]
\centering
\small
\resizebox{2.0\columnwidth}{!}{
\begin{tabular}{c ccc cc ccc}
\toprule
              & \multicolumn{3}{c}{Schema ranking} & \multicolumn{2}{c}{Document ranking} & \multicolumn{3}{c}{Corpus coverage} \\ \cmidrule(r){2-4} \cmidrule(r){5-6} \cmidrule(r){7-9}
$\mathrm{N}_{\mathrm{events}}$  & Avg Rank$\downarrow$                  & MRR$\uparrow$                       & R@10$\uparrow$                      & R@30$\uparrow$                      & nDCG$\uparrow$                      & Cov@0.5$\uparrow$           & Cov@0.7$\uparrow$                   & Cov@0.9$\uparrow$ \\ \midrule
$[1;5)$                         & 26.4 \myspace \emph{35.4} & .112 \myspace \emph{.072} & .244 \myspace \emph{.199} & .387 \myspace \emph{.293} & .246 \myspace \emph{.162} & .960 \myspace \emph{.895} & .852 \myspace \emph{.576} & .797 \myspace \emph{.491} \\
$[5;10)$                        & 23.8 \myspace \emph{32.1} & .147 \myspace \emph{.088} & .340 \myspace \emph{.193} & .472 \myspace \emph{.347} & .276 \myspace \emph{.170} & .937 \myspace \emph{.833}  & .785 \myspace \emph{.502}  & .614 \myspace \emph{.334}     \\
$[10; \infty)$                  & 20.8 \myspace \emph{30.6}  & .194 \myspace \emph{.105}  & .410 \myspace \emph{.229}  & .545 \myspace \emph{.411}  & .269 \myspace \emph{.247}  & .925 \myspace \emph{.759}  & .759 \myspace \emph{.417}  & .533 \myspace \emph{.242}    \\ \addlinespace
$[1; \infty)$                   & 21.1 \myspace \emph{30.2}  & .191 \myspace \emph{.109}  & .404 \myspace \emph{.239}  & .442 \myspace \emph{.351}  & .272 \myspace \emph{.240}  & .925 \myspace \emph{.745}  & .761 \myspace \emph{.400}  & .542 \myspace \emph{.223}  \\
\bottomrule
\end{tabular}}
\caption{Summary of the ranking-based evaluation over 82 schemas and documents from \textbf{KAIROS SLC}. Numbers in regular font use gold events from the corpus, numbers in \emph{italics} use events extracted with the LOME IE system.}\label{tbl:ldc}
\end{table*}

\begin{table}[t!]
\centering
\small
\resizebox{0.9\columnwidth}{!}{
\begin{tabular}{c ccc ccc}
\toprule
              & \multicolumn{3}{c}{82 schemas} & \multicolumn{3}{c}{232 schemas} \\ \cmidrule(r){2-4} \cmidrule(r){5-7}
$\mathrm{N}_{\mathrm{events}}$ & 0.5 & 0.7 & 0.9 & 0.5 & 0.7 & 0.9  \\ \midrule
$[1;5)$        & .887 & .531 & .425 & .975 & .637 & .509      \\
$[5;10)$       & .791 & .391 & .233 & .892 & .496 & .278      \\
$[10; \infty)$ & .695 & .313 & .164 & .807 & .379 & .195     \\ \addlinespace
$[1; \infty)$  & .684 & .303 & .154 & .798 & .367 & .183  \\ 
\bottomrule
\end{tabular}}
\caption{Corpus coverage Cov@$t$ (${t \in \{0.5, 0.7, 0.9\}}$) on the \textbf{Gigaword} corpus, using events extracted with the LOME IE system.}\label{tbl:gw-lome}
\end{table}

\begin{table}[t!]
\centering
\small
\resizebox{0.9\columnwidth}{!}{
\begin{tabular}{c ccc ccc}
\toprule
              & \multicolumn{3}{c}{82 schemas} & \multicolumn{3}{c}{232 schemas} \\ \cmidrule(r){2-4} \cmidrule(r){5-7}
$\mathrm{N}_{\mathrm{events}}$ & 0.5 & 0.7 & 0.9 & 0.5 & 0.7 & 0.9  \\ \midrule
$[1;5)$        & .874 & .588 & .529 & .980 & .719 & .643 \\
$[5;10)$       & .784 & .450 & .303 & .915 & .558 & .368 \\
$[10; \infty)$ & .708 & .376 & .224 & .850 & .472 & .272 \\ \addlinespace
$[1; \infty)$  & .720 & .392 & .246 & .860 & .490 & .299  \\
\bottomrule
\end{tabular}}
\caption{Corpus coverage Cov@$t$ (${t \in \{0.5, 0.7, 0.9\}}$) on the \textbf{English} subset of the \textbf{CC-News} corpus, using events extracted with the LOME IE system.}\label{tbl:clir-lome}
\end{table}

Event schemas are meant to provide missing pieces of knowledge (e.g., events and their participants) that are otherwise not stated explicitly in text, aiding document-level tasks such as coreference, summarization, and inference \cite{chambers2010database,balasubramanian2013generating}. When dealing with a large schema library $L$, one needs to first narrow down all schemas $s \in L$ to only those that apply to a given document $d$, depending on the task. We quantify this with a similarity function $sim(d, s)$ and a task-specific threshold $t$: namely, we say that $s$ applies to $d$ when $sim(d, s) \geq t$ for some task-specific $t$. Given $t$, we compute \emph{coverage at} $t$ (Cov@$t$) as a fraction of documents $d \in D$ such that at least one schema $s \in L$ applies to $d$:
$$
\mathrm{Cov@}t = \frac{|\{d \in D \mid \exists s \in L: sim(d, s) \geq t \}|}{|D|}.
$$

We compute Cov$@t$ for the 82 schema subset, and for the full 232 schemas' library.
We use  ${sim(d, s)= |E(d) \cap E(s)|/|E(d)|}$, where $E(s)$ and $E(d)$ define all events mentioned in a schema $s$ and extracted from a document $d$, respectively.\footnote{For our experiments, we treat both $E(s)$ and $E(d)$ as multisets of events.
E.g., if a document $d$ such that ${E(d) = \{\textsc{Life.Infect}, \textsc{Life.Infect}, \textsc{Medical.Vaccinate}\}}$ is matched with a schema $s$ such that ${E(s) = \{\textsc{Life.Infect}, \textsc{Life.Die}\}}$, then ${sim(d, s) = 2/3}$.}
The events are automatically extracted using the pretrained multilingual FrameNet parser from the LOME IE system \cite{xia2021lome}, which extracts FrameNet events and their arguments. To account for varying document lengths, we stratify the results by the number of extracted events $\mathrm{N}_{\mathrm{events}}$ in each document. We map the extracted FrameNet events to the KAIROS ontology using a rule-based mapping.\footnote{The mapping rules can be accessed at this link:\\ \url{https://nlp.jhu.edu/schemas/k2f.js}}

As a result, we observe (\Cref{tbl:gw-lome,tbl:clir-lome}) that the initial 82 schemas cover a meaningful part of Gigaword and CC-News: at least 15-25\%, and up to 98.3\% of documents, depending on corpus $D$ and threshold $t$. Extending the library $L$ by the additional 150 schemas improves corpus coverage by around 20\%, thus suggesting these 150 schemas improve the diversity of the scenarios covered by the initial 82 schemas.

\begin{table}[t!]
\centering
\small
\resizebox{0.9\columnwidth}{!}{
\begin{tabular}{c ccc ccc}
\toprule
              & \multicolumn{3}{c}{82 schemas} & \multicolumn{3}{c}{232 schemas} \\ \cmidrule(r){2-4} \cmidrule(r){5-7}
$\mathrm{N}_{\mathrm{events}}$ & 0.5 & 0.7 & 0.9 & 0.5 & 0.7 & 0.9  \\ \midrule
$[1;5)$        & .886 & .612 & .561 & .983 & .734 & .670 \\
$[5;10)$       & .778 & .465 & .335 & .921 & .586 & .408 \\
$[10; \infty)$ & .688 & .387 & .250 & .839 & .492 & .306 \\ \addlinespace
$[1; \infty)$  & .713 & .414 & .287 & .858 & .523 & .349   \\
\bottomrule
\end{tabular}}
\caption{Corpus coverage Cov@$t$ (${t \in \{0.5, 0.7, 0.9\}}$) on the \textbf{Russian} subset of the \textbf{CC-News} corpus, using events extracted with the LOME IE system.}\label{tbl:clir-ru-lome}
\end{table}

\begin{table}[t!]
\centering
\small
\resizebox{0.9\columnwidth}{!}{
\begin{tabular}{c ccc ccc}
\toprule
              & \multicolumn{3}{c}{82 schemas} & \multicolumn{3}{c}{232 schemas} \\ \cmidrule(r){2-4} \cmidrule(r){5-7}
$\mathrm{N}_{\mathrm{events}}$ & 0.5 & 0.7 & 0.9 & 0.5 & 0.7 & 0.9  \\ \midrule
$[1;5)$        & .875 & .589 & .528 & .981 & .718 & .639 \\
$[5;10)$       & .776 & .460 & .314 & .924 & .582 & .387 \\
$[10; \infty)$ & .699 & .408 & .251 & .877 & .531 & .314 \\ \addlinespace
$[1; \infty)$  & .713 & .422 & .271 & .885 & .545 & .337 \\
\bottomrule
\end{tabular}}
\caption{Corpus coverage Cov@$t$ (${t \in \{0.5, 0.7, 0.9\}}$) on the \textbf{Chinese} subset of the \textbf{CC-News} corpus, using events extracted with the LOME IE system.}\label{tbl:clir-zh-lome}
\end{table}

Comparing across multiple languages in CC-News (\Cref{tbl:clir-lome,tbl:clir-ru-lome,tbl:clir-zh-lome}), we notice the coverage on Chinese and Russian news articles does not drop and even improves, despite that schemas were originally mined using English-language resources. This suggests that the proposed schemas are robust and useful for cross-lingual scenarios, owing to its language-independent ontology and the advances in cross-lingual event extraction tools.

The difference between 82 and 232 schemas' coverage is significant (p-value~${\ll 0.01}$) for all compared variations of $N_{events}$ and $t$, as measured by the two-sided Wilcoxon signed-rank test.

\subsection{Ranking Evaluation}\label{app:ranking_eval}
How sufficient is the event-only representation $E(d)$ of a document $d$ to rank schemas ${s \in L}$ and predict the true complex event (CE), using ${sim(d, s)}$ as a ranking function? To answer this question, we conduct a ranking evaluation using KAIROS SLC, where each $d$ has precisely one CE label. For each document $d$, we \textbf{rank schemas} according to ${sim(d, s)}$ and report the average rank (lower is better), mean reciprocal rank (MRR, higher is better), and recall@10 (R@10, higher is better) of the gold CE label. Similarly we ask, how well can we rank schema-salient documents ${d \in D}$ given event-only description $E(s)$ of a schema $s$? For each schema $s$, we \textbf{rank documents} according to ${sim(d, s)}$ and report recall@30 (higher is better) and normalized discounted cumulative gain (nDCG, higher is better) of the gold annotated documents. We also compute \textbf{corpus coverage}, which does not require ground-truth CE labels.

As a result (\Cref{tbl:ldc}), we find that the event-only representation does provide useful signal for ranking documents and schemas, compared to e.g. a fully random ordering (where R@10 for schema ranking ${= \frac{10}{82} \approx 0.122}$ and R@30 for document ranking ${= \frac{30}{921} \approx 0.033}$). Including additional signal, like participants' types and relations, could potentially improve the ranking. However, this information is costly to annotate for, and was not provided for most of the documents in KAIROS SLC. Thus, improving annotation pipelines for complex events could not only boost schema induction, as argued throughout our paper, but also enable rapid data collection for schema-based information extraction, which in turn leads to more precise schema-supported inferences in downstream document-level tasks.

\section{Conclusions}
In this paper, we propose a novel semi-automatic script induction system and  induce a dataset of 232 schemas. The automatic portion of our system is rooted in a new method, extending an ARM-based approach, which finds interesting subsequences, with a causal scoring metric for filtering out and fusing together these interesting subsequences. The interactive portion of our system is made possible through a new tool, SchemaBlocks, a block-based interface developed to make annotation of schema structures intuitive and easy.

We release both the SchemaBlocks interface and the induced 232 schemas to the community, which we believe will be useful broadly and will facilitate further efforts in what is traditionally an interminable pain for all looking to build robust AI systems: the annotation of robust commonsense knowledge structures.

\section{Bibliographical References}\label{reference}
\bibliographystyle{lrec2022-bib}
\bibliography{paper}

\begin{thebibliography}{}

\bibitem[\protect\citename{Abbott \bgroup et al.\egroup
  }1985]{abbott1985representation}
Abbott, V., Black, J.~B., and Smith, E.~E.
\newblock (1985).
\newblock The representation of scripts in memory.
\newblock {\em Journal of memory and language}, 24(2):179--199.

\bibitem[\protect\citename{Bach \bgroup et al.\egroup }2017]{bach2017hinge}
Bach, S.~H., Broecheler, M., Huang, B., and Getoor, L.
\newblock (2017).
\newblock Hinge-loss {M}arkov random fields and probabilistic soft logic.
\newblock {\em Journal of Machine Learning Research (JMLR)}.

\bibitem[\protect\citename{Baker \bgroup et al.\egroup
  }1998]{baker1998berkeley}
Baker, C.~F., Fillmore, C.~J., and Lowe, J.~B.
\newblock (1998).
\newblock The berkeley framenet project.
\newblock In Christian Boitet et~al., editors, {\em 36th Annual Meeting of the
  Association for Computational Linguistics and 17th International Conference
  on Computational Linguistics, {COLING-ACL} '98, August 10-14, 1998,
  Universit{\'{e}} de Montr{\'{e}}al, Montr{\'{e}}al, Quebec, Canada.
  Proceedings of the Conference}, pages 86--90.

\bibitem[\protect\citename{Balasubramanian \bgroup et al.\egroup
  }2013]{balasubramanian2013generating}
Balasubramanian, N., Soderland, S., Mausam, O.~E., and Etzioni, O.
\newblock (2013).
\newblock Generating coherent event schemas at scale.
\newblock In {\em Proceedings of EMNLP}.

\bibitem[\protect\citename{Belyy and
  Van~Durme}2020]{belyy-van-durme-2020-script}
Belyy, A. and Van~Durme, B.
\newblock (2020).
\newblock Script induction as association rule mining.
\newblock In {\em Proceedings of the First Joint Workshop on Narrative
  Understanding, Storylines, and Events}, pages 55--62, Online, July.
  Association for Computational Linguistics.

\bibitem[\protect\citename{Bower \bgroup et al.\egroup }1979]{bower1979scripts}
Bower, G.~H., Black, J.~B., and Turner, T.~J.
\newblock (1979).
\newblock Scripts in memory for text.
\newblock {\em Cognitive psychology}, 11(2):177--220.

\bibitem[\protect\citename{Chambers and Jurafsky}2008]{chambers:08}
Chambers, N. and Jurafsky, D.
\newblock (2008).
\newblock Unsupervised learning of narrative event chains.
\newblock In {\em Proceedings of the Association for Computational Linguistics
  (ACL)}, Hawaii, USA.

\bibitem[\protect\citename{Chambers and Jurafsky}2009]{chambers:09}
Chambers, N. and Jurafsky, D.
\newblock (2009).
\newblock Unsupervised learning of narrative schemas and their participants.
\newblock In {\em Proceedings of the Association for Computational Linguistics
  (ACL)}, Singapore.

\bibitem[\protect\citename{Chambers and Jurafsky}2010]{chambers2010database}
Chambers, N. and Jurafsky, D.
\newblock (2010).
\newblock A database of narrative schemas.
\newblock In {\em LREC}.

\bibitem[\protect\citename{Chang \bgroup et al.\egroup }2009]{chang09reading}
Chang, J., Boyd{-}Graber, J.~L., Gerrish, S., Wang, C., and Blei, D.~M.
\newblock (2009).
\newblock Reading tea leaves: How humans interpret topic models.
\newblock In Yoshua Bengio, et~al., editors, {\em Advances in Neural
  Information Processing Systems 22: 23rd Annual Conference on Neural
  Information Processing Systems 2009. Proceedings of a meeting held 7-10
  December 2009, Vancouver, British Columbia, Canada}, pages 288--296. Curran
  Associates, Inc.

\bibitem[\protect\citename{DeJong}1983]{dejong1983acquiring}
DeJong, G.
\newblock (1983).
\newblock Acquiring schemata through understanding and generalizing plans.
\newblock In {\em IJCAI}.

\bibitem[\protect\citename{Doddington \bgroup et al.\egroup
  }2004]{doddington2004automatic}
Doddington, G.~R., Mitchell, A., Przybocki, M.~A., Ramshaw, L.~A., Strassel,
  S.~M., and Weischedel, R.~M.
\newblock (2004).
\newblock The automatic content extraction (ace) program-tasks, data, and
  evaluation.
\newblock In {\em Lrec}, volume~2, pages 837--840. Lisbon.

\bibitem[\protect\citename{Graff \bgroup et al.\egroup }2003]{graff2003english}
Graff, D., Kong, J., Chen, K., and Maeda, K.
\newblock (2003).
\newblock English gigaword.
\newblock {\em Linguistic Data Consortium, Philadelphia}, 4(1):34.

\bibitem[\protect\citename{Han \bgroup et al.\egroup }2000]{han2000mining}
Han, J., Pei, J., and Yin, Y.
\newblock (2000).
\newblock Mining frequent patterns without candidate generation.
\newblock {\em ACM sigmod record}, 29(2):1--12.

\bibitem[\protect\citename{Kiros \bgroup et al.\egroup }2015]{kiros15skip}
Kiros, R., Zhu, Y., Salakhutdinov, R., Zemel, R.~S., Urtasun, R., Torralba, A.,
  and Fidler, S.
\newblock (2015).
\newblock Skip-thought vectors.
\newblock In Corinna Cortes, et~al., editors, {\em Advances in Neural
  Information Processing Systems 28: Annual Conference on Neural Information
  Processing Systems 2015, December 7-12, 2015, Montreal, Quebec, Canada},
  pages 3294--3302.

\bibitem[\protect\citename{Lehnert \bgroup et al.\egroup }1983]{lehnert83boris}
Lehnert, W.~G., Dyer, M.~G., Johnson, P.~N., Yang, C.~J., and Harley, S.
\newblock (1983).
\newblock {BORIS} - an experiment in in-depth understanding of narratives.
\newblock {\em Artif. Intell.}, 20(1):15--62.

\bibitem[\protect\citename{Minsky}1974]{minsky74}
Minsky, M.
\newblock (1974).
\newblock A framework for representing knowledge.
\newblock MIT Laboratory Memo 306.

\bibitem[\protect\citename{Mooney and DeJong}1985]{mooney}
Mooney, R. and DeJong, G.
\newblock (1985).
\newblock Learning schemata for natural language processing.
\newblock In {\em Ninth International Joint Conference on Artificial
  Intelligence (IJCAI)}, pages 681--687.

\bibitem[\protect\citename{Mueller}1999]{mueller1999database}
Mueller, E.~T.
\newblock (1999).
\newblock A database and lexicon of scripts for thoughttreasure.

\bibitem[\protect\citename{Nagel}2016]{nagel2016cc}
Nagel, S.
\newblock (2016).
\newblock Cc-news.
\newblock \url{https://commoncrawl.org/2016/10/news-dataset-available}.

\bibitem[\protect\citename{Regneri \bgroup et al.\egroup
  }2010]{regneri2010learning}
Regneri, M., Koller, A., and Pinkal, M.
\newblock (2010).
\newblock Learning script knowledge with web experiments.
\newblock In {\em Proceedings of the 48th Annual Meeting of the Association for
  Computational Linguistics}, pages 979--988. Association for Computational
  Linguistics.

\bibitem[\protect\citename{Resnick \bgroup et al.\egroup
  }2009]{resnick2009scratch}
Resnick, M., Maloney, J., Monroy-Hern{\'a}ndez, A., Rusk, N., Eastmond, E.,
  Brennan, K., Millner, A., Rosenbaum, E., Silver, J., Silverman, B., et~al.
\newblock (2009).
\newblock Scratch: programming for all.
\newblock {\em Communications of the ACM}, 52(11):60--67.

\bibitem[\protect\citename{Rudinger \bgroup et al.\egroup }2015]{Rudinger2015}
Rudinger, R., Rastogi, P., Ferraro, F., and Van~Durme, B.
\newblock (2015).
\newblock Script induction as language modeling.
\newblock In {\em Proceedings of the 2015 Conference on Empirical Methods in
  Natural Language Processing}, pages 1681--1686.

\bibitem[\protect\citename{Schank and Abelson}1977]{sch77scripts}
Schank, R.~C. and Abelson, R.~P.
\newblock (1977).
\newblock {\em Scripts, plans, goals, and understanding: An inquiry into human
  knowledge structures}.
\newblock Psychology Press.

\bibitem[\protect\citename{Song \bgroup et al.\egroup }2015]{song2015light}
Song, Z., Bies, A., Strassel, S., Riese, T., Mott, J., Ellis, J., Wright, J.,
  Kulick, S., Ryant, N., and Ma, X.
\newblock (2015).
\newblock From light to rich ere: annotation of entities, relations, and
  events.
\newblock In {\em Proceedings of the the 3rd Workshop on EVENTS: Definition,
  Detection, Coreference, and Representation}, pages 89--98.

\bibitem[\protect\citename{Wanzare \bgroup et al.\egroup }2016]{descript}
Wanzare, L. D.~A., Zarcone, A., Thater, S., and Pinkal, M.
\newblock (2016).
\newblock A crowdsourced database of event sequence descriptions for the
  acquisition of high-quality script knowledge.
\newblock In {\em Proceedings of the Tenth International Conference on Language
  Resources and Evaluation ({LREC}'16)}, pages 3494--3501, Portoro{\v{z}},
  Slovenia, May. European Language Resources Association (ELRA).

\bibitem[\protect\citename{Wanzare \bgroup et al.\egroup
  }2017]{descript_clusters}
Wanzare, L., Zarcone, A., Thater, S., and Pinkal, M.
\newblock (2017).
\newblock Inducing script structure from crowdsourced event descriptions via
  semi-supervised clustering.
\newblock In {\em Proceedings of the 2nd Workshop on Linking Models of Lexical,
  Sentential and Discourse-level Semantics}, pages 1--11, Valencia, Spain,
  April. Association for Computational Linguistics.

\bibitem[\protect\citename{Weber \bgroup et al.\egroup
  }2020]{weber-etal-2020-causal}
Weber, N., Rudinger, R., and Van~Durme, B.
\newblock (2020).
\newblock Causal inference of script knowledge.
\newblock In {\em Proceedings of the 2020 Conference on Empirical Methods in
  Natural Language Processing (EMNLP)}, pages 7583--7596, Online, November.
  Association for Computational Linguistics.

\bibitem[\protect\citename{Xia \bgroup et al.\egroup }2021]{xia2021lome}
Xia, P., Qin, G., Vashishtha, S., Chen, Y., Chen, T., May, C., Harman, C.,
  Rawlins, K., White, A.~S., and Durme, B.~V.
\newblock (2021).
\newblock {LOME:} large ontology multilingual extraction.
\newblock In Dimitra Gkatzia et~al., editors, {\em Proceedings of the 16th
  Conference of the European Chapter of the Association for Computational
  Linguistics: System Demonstrations, {EACL} 2021, Online, April 19-23, 2021},
  pages 149--159. Association for Computational Linguistics.

\bibitem[\protect\citename{Zhu \bgroup et al.\egroup }2015]{zhu15aligning}
Zhu, Y., Kiros, R., Zemel, R.~S., Salakhutdinov, R., Urtasun, R., Torralba, A.,
  and Fidler, S.
\newblock (2015).
\newblock Aligning books and movies: Towards story-like visual explanations by
  watching movies and reading books.
\newblock In {\em 2015 {IEEE} International Conference on Computer Vision,
  {ICCV} 2015, Santiago, Chile, December 7-13, 2015}, pages 19--27. {IEEE}
  Computer Society.

\end{thebibliography}

\clearpage
\appendix
\section{Schema Intrusion Task Details}

\subsection{Schema Inference}
\label{app:schema_inference}

Here, we describe how we match schemas with documents in the schema intrusion task. There are 3 main parts to the Schema Inference system:
\begin{enumerate}
    \item representations for events and participants,
    \item representations for schemas, and
    \item the inference mechanism based on Probabilistic Soft Logic (PSL) \cite{bach2017hinge}.
\end{enumerate}

Throughout the following, we will use the example depicted in \Cref{fig:schema}.

\paragraph*{Events and participants}

Each document is turned into a knowledge graph using a FrameNet parser, as described in \Cref{sec:eval}. Knowledge graphs are then flattened to unary or binary relations, following neo-Davidsonian semantics. For instance,

\begin{scriptsize}
\begin{verbatim}
{ "@id": "K0C03N60D.7.2",
  "@type": "kairos:Primitives/Events/
            Movement.Transportation.Unspecified",
  "confidence": 0.9,
  "participants": [
    { "@id": "K0C03N60D.7.2.P1.1",
      "role": "kairos:Primitives/Events/
               Movement.Transportation.Unspecified/
               Slots/Destination",
      "values": [{ "confidence": 1.0,
                   "entity": "e2323a3", }]},  
    { "@id": "K0C03N60D.7.2.P3.1",
      "role": "kairos:Primitives/Events/
               Movement.Transportation.Unspecified/
               Slots/PassengerArtifact",
      "values": [{ "confidence": 0.8,
                   "entity": "e2323a1", }]}
  ],
}
\end{verbatim}
\end{scriptsize}

is turned into

\begin{scriptsize}
\begin{verbatim}

Movement.Transportation.Unspecified(K0C03N60D.7.2) .9

Destination(K0C03N60D.7.2, e2323a3) 1.

PassengerArtifact(K0C03N60D.7.2, e2323a1) .8

\end{verbatim}
\end{scriptsize}

We omit common prefixes for readability. We collect those predicates in dedicated files, together with confidence values, which constitute PSL's observation files.

\paragraph*{Schemas}

We frame each step in a schema as a predicate, whose arguments are an event and a number of participants. We frame a schema as a predicate, whose arguments are a set of events, where each is an argument to one of its steps. Using Horn clauses, we define the schema as a conjunction of its steps.

Concretely, the example from \Cref{fig:schema} turns into:

\begin{scriptsize}
\begin{verbatim}
Territorial_Dispute(Claim_event, Attack_event,
        Diplomatic_event)
    <- Claim(Claim_event, State1, State2, Territory)
    & Attack(Attack_event, State1,
        State2, Territory)
    & Diplomatic_Resolution(Diplomatic_event,
        State1, State2, Territory)

Territorial_Dispute(Claim_event, Attack_event,
        Resolution_event)
    <- Claim(Claim_event, State1, State2, Territory)
    & Attack(Attack_event, State1,
        State2, Territory)
    & Legal_Resolution(Resolution_event, State1,
        State2, InternationalCourt, Territory)
        
Claim(Contact_event, State1, State2, Territory)
    <- Contact.Contact.Unspecified(Contact_event)
    & Participant(Contact_event, State1)
    & Participant(Contact_event, State2)
    & Topic(Contact_event, Territory)

Attack(Attack_event, State1, State2, Territory)
    <- Conflict.Attack.Unspecified(Attack_event)
    & Attacker(Attack_event, State1)
    & Target(Attack_event, State2)
    & Place(Attack_event, Territory)

Diplomatic_Resolution(Contact_event, State1,
        State2, Territory)
    <- Contact.Contact.Unspecified(Contact_event)
    & Participant(Contact_event, State1)
    & Participant(Contact_event, State2)
    & Topic(Contact_event, Territory)

Legal_Resolution(Contact_event, State1, State2,
        InternationalCourt, Territory)
    <- Contact.Contact.Unspecified(Contact_event)
    & Participant(Contact_event, State1)
    & Participant(Contact_event, State2)
    & Participant(Contact_event, InternationalCourt)
    & Topic(Contact_event, Territory)
\end{verbatim}
\end{scriptsize}

We include negative priors for each step and schema predicate. We give each rule a weight: $100$ for step definitions, $10$ for schema definitions, and $1$ for negative priors. The negative priors and the weights jointly ensure that with a rule of the form \texttt{A \& B -> C} where \texttt{A} and \texttt{B} are ground expressions, \texttt{C} will be assigned the probability assigned to \texttt{A \& B}. Primitive events from the ontology and typing predicates are set to closed predicates. Other predicates are set to open.

\paragraph*{PSL Inference}

PSL is a formalism and a tool to assign probabilities to ground expressions.
To perform schema inference, we enumerate all the possible groundings for schemas and steps, i.e. all possible combinations of predicates and arguments. The set of possible arguments is taken as the set of entities and events from the knowledge graph. The set of possible predicates is that of all possible events and slots from the KAIROS ontology. PSL associates a continuous variable to each of those targets, and uses the observation files and the rule file to produce a convex optimization problem involving those variables. Solving this optimization problem results in values for the variables, which we interpret as the probability, for individual events, steps and schemas, that they have happened. To be able to partially match a schema, we need to be able to ground any subset of its events and participants. We do this by introducing ``UNK'' events and entities, which can fill any event and participant role, and co-refer with any entity.

We post-process PSL's results to obtain instantiated schemas, using the confidence values provided by PSL. Any event whose value is an ``UNK'' event, we consider as unmatched, and interpret this as an event that was not found in the documents, but that is predicted by the schema to have happened. We re-scale the confidence of a schema by the proportion of matched events it contains.

To simplify the matching process, we filter the schema library using an Apache Lucene index. Schemas and knowledge graphs are represented as bags-of-events. We build a Lucene index for the schema library, and given a knowledge graph, query it for relevant schemas.

\subsection{Human Evaluation}\label{app:eval_details}
\begin{figure*}[b]
\centering
\vspace{0.5cm}
\includegraphics[width=1.00\textwidth]{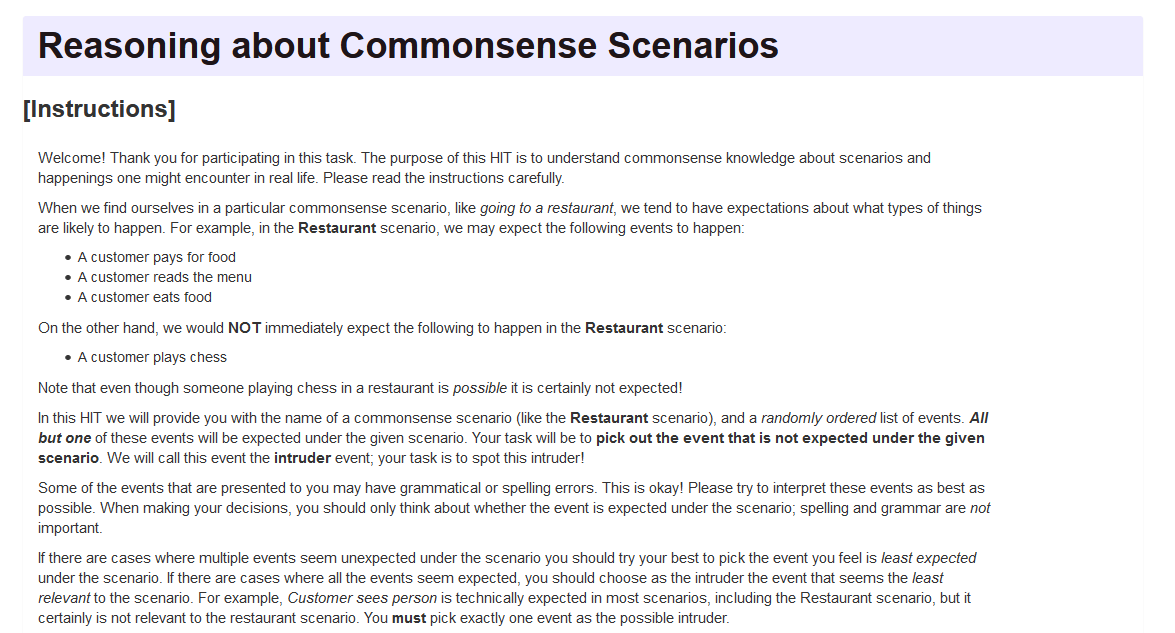}
\caption{Instructions for the schema intrusion task shown to the Amazon Mechanical Turk workers.}
\vspace{0.8cm}
\label{fig:intruder_instructions}
\end{figure*}

\begin{figure*}[b]
\centering
\includegraphics[width=1.00\textwidth]{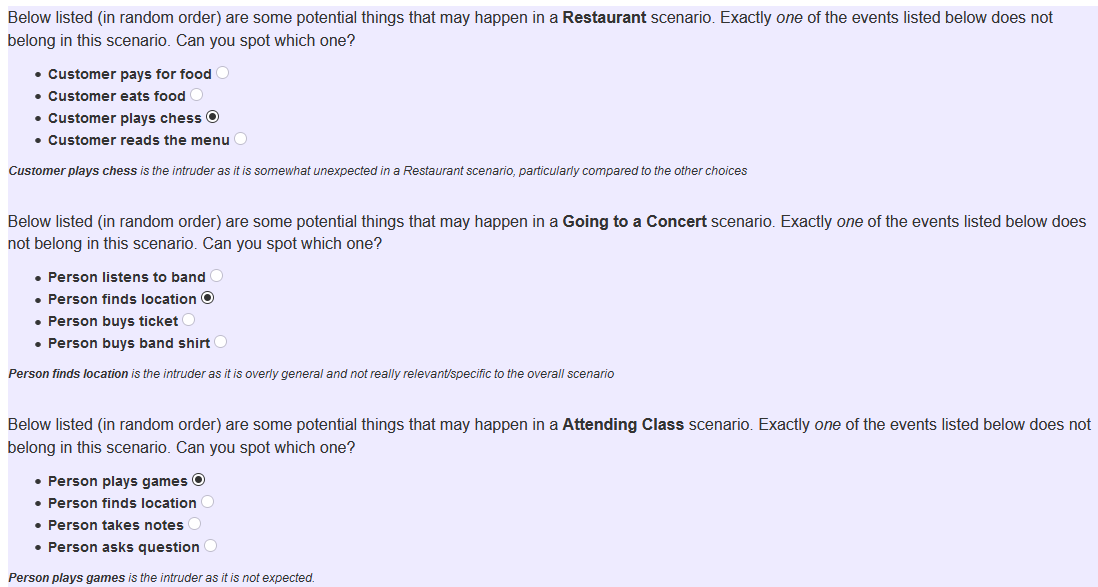}
\caption{Examples of schemas along with intruder events shown to the Amazon Mechanical Turk workers.}
\label{fig:intruder_examples}
\end{figure*}

We use Mechanical Turk to collect responses for the schema intrusion task. Each Mechanical Turk assignment consists of a \emph{single} intrusion task (i.e. a single schema with an intruder, see \Cref{fig:intruder}(b)). Each task is completed by three separate annotators who are paid \$0.20 per assignment. Instructions shown to the annotators can be seen on \Cref{fig:intruder_instructions,fig:intruder_examples}.

\end{document}